\documentclass[10pt,twocolumn,letterpaper]{article}

\usepackage[pagenumbers]{iccv} 

\definecolor{iccvblue}{rgb}{0.21,0.49,0.74}
\usepackage[pagebackref,breaklinks,colorlinks,allcolors=iccvblue]{hyperref}
\usepackage{amssymb}
\usepackage{booktabs}
\usepackage{multirow}

\def\paperID{14040} 
\def\confName{ICCV}
\def\confYear{2025}

\title{DVD: A Comprehensive Dataset for Advancing Violence Detection in Real-World Scenarios}

\author{Dimitrios Kollias$^{1,2}$, \hspace{0.3cm}
Damith C. Senadeera$^{1,2}$, \hspace{0.3cm}
Jianian Zheng$^{1}$, \hspace{0.3cm}
Kaushal K. K. Yadav$^{1}$,\\
Greg Slabaugh$^{1,2}$, \hspace{0.3cm}
Muhammad Awais$^{1,2}$, \hspace{0.3cm}
Xiaoyun Yang2$^{3}$\\ \\
$^{1}$Queen Mary University of London, UK \\
$^{2}$Queen Mary’s Digital Environment Research Institute (DERI), UK \\
$^{3}$Remark AI UK Ltd, UK \\
}

\begin{document}
\maketitle
\begin{abstract}
Violence Detection (VD) has become an increasingly vital area of research.
Existing automated VD efforts are hindered by the limited availability of diverse, well-annotated databases. 
Existing databases suffer from coarse video-level annotations, limited scale and diversity, and lack of metadata, restricting the generalization of models.
To address these challenges, we introduce DVD, a large-scale (500 videos, 2.7M frames), frame-level annotated VD database with diverse environments, varying lighting conditions, multiple camera sources, complex social interactions, and rich metadata. DVD is designed to capture the complexities of real-world violent events. 
\end{abstract}

\section{Introduction}
\label{sec:intro}

Violence is a pervasive issue in modern society, with severe consequences for individuals and communities alike. From urban confrontations to domestic disputes, the ability to detect and respond to violent events in real time is critical for ensuring public safety and mitigating harm. Manual surveillance methods, however, are limited in their scalability and efficacy, particularly in high-density or high-risk areas. Manual monitoring of potential violent activities is labor-intensive and prone to human error. Therefore there is a critical need for automated violence detection systems that leverage the advancements in artificial intelligence and deep learning to identify and analyze violent incidents efficiently and accurately.

Violence can be broadly categorized into: i) \textit{action violence}, characterized by physical activities (e.g., fighting, assaults), which can be classified into individual violence and crowd violence; and ii) \textit{hybrid violence}, which combines actions with other indicators (e.g., verbal aggression, environmental disturbances, presence of weapons, explosions). Despite the growing interest in automated violence detection, the development of robust models is hampered by the lack of large-scale, diverse, and representative databases. 
Current databases present numerous limitations that constrain the training and evaluation of advanced models.

The challenges of violence detection are manifold. Violent scenes are inherently diverse, taking place in a wide range of settings (e.g., indoors/outdoors), under varying lighting conditions (e.g., day/night), and captured by different types of cameras (e.g., surveillance cameras, mobile phones, body cameras). Additional challenges include occlusion, crowd density, resolution variability, and the presence of environmental noise, all of which complicate the detection process. Addressing these challenges requires not only advanced algorithms but also large-scale databases that reflect the complexity and variability of real-world violence.

\begin{table*}[t]
\centering
\caption{Summary of video violence databases. `V' denotes violent frames, while `NV' represents non-violent frames.}
\scalebox{0.8}{
\begin{tabular}{lccccccccc}
\toprule
\textbf{Database} & \textbf{Year} & \textbf{Scale} & \textbf{\# Frames V / NV}  & \textbf{Length (sec)} & \textbf{Resolution} & \textbf{Annotation} & \textbf{Scenario} & \textbf{Type}  & \textbf{Access}    \\
\toprule

Movies~\cite{movie_fight} & 2011 & 200 Clips & 100 / 100 & 1.6-2   & $720 \times 480$ & Video-Level & Movie & Action & Public \\
\hline

Crowd Violence~\cite{crowd_violence} & 2012 & 246 Clips & 123 / 123 & 1.04-6.52   & Variable & Video-Level & Natural & Action & Public   \\
\hline

Hockey Fight~\cite{hockeyfight} & 2012 & 1,000 Clips & 500 / 500  & 1.6-1.96& $360 \times 288$ & Video-Level & Hockey Games & Action & Public \\
\hline

SBU Kinect ~\cite{sbu_kinetics} & 2012 & 264 Clips & - & 0.67-3 & $640 \times 480$ & Video-Level & Acted Fights & Action & Public \\
\hline


XD Violence~\cite{xdviolence} & 2020 & 4750 Clips & 2405 / 2349 &10-600+    & Variable & Video-Level & Variable & Hybrid & Public  \\
\hline

UCF Crime~\cite{ucfcrime_dataset} & 2018 & 1900 Clips & 950 / 950 &60-600    & Variable & Video-Level & Surveillance & Hybrid & Public  \\
\hline

RLVS~\cite{rlvs} & 2019 & 2000 Clips & 1000 / 1000 &3-7    & Variable & Video-Level & YouTube & Action & Public  \\
\hline

Human Violence & 2019 & 1930 Clips & 1930 / 0 & - & $1920 \times 1080$ & Video-Level  &   Promotion  & Hybrid & Private  \\
\hline

RWF-2000~\cite{rwf_dataset} & 2020 & 2,000 Clips & 1000 / 1000 & 5 & Variable & Video-Level & Surveillance & Action & Public \\
\hline

VioShot~\cite{xiang2024}& 2024 & 2500 Clips & 2000 / 500 & 0.14-40.53 &  $1920 \times 1080$ & Video-Level  &  Movie  & Hybrid  & Public \\
\hline

VioPeru~\cite{baca2024} & 2024 & 367 Clips & 280 / 87 & 5 & Variable & Video-Level & Surveillance &  Hybrid  & Public \\
\bottomrule

\textbf{DVD (Ours)} & \textbf{2025} & \begin{tabular}{@{}c@{}} \textbf{500 Videos} \\ \textbf{2.7M Frames}\end{tabular} & \begin{tabular}{@{}c@{}}  \textbf{900K / 1.6M} \end{tabular} & \textbf{15-1200} & \textbf{Variable} & \textbf{Frame-Level} & \textbf{Variable} & \textbf{Hybrid} & \textbf{Public}  \\
\bottomrule

\end{tabular}
}
\label{dbs}
\end{table*}

Existing violence detection databases suffer from several critical shortcomings, with \underline{Annotation Granularity} being one. Most widely used databases provide video-level annotations (as shown in Table \ref{dbs}), where an entire video is labeled as either violent or non-violent. This coarse labeling approach overlooks the temporal dynamics of violence, failing to account for instances where both violent and non-violent events coexist within the same video.
%
%
Moreover, video-level annotations fail to capture variations in intensity, duration, and context of violent incidents. For instance, a brief yet significant altercation in a lengthy video may carry critical importance, but its impact is diluted by the surrounding non-violent content, resulting in noisy  data. Assigning a single label to such mixed-content videos introduces ambiguity and inconsistency in annotations, further hindering model performance.
Prominent databases like VioShot~\cite{xiang2024}, VioPeru~\cite{baca2024}, RWF-2000~\cite{rwf_dataset}   rely on video-level annotations, limiting their granularity and utility.

The second shortcoming is the \underline{Limited Scale and Diversity}. Many databases are small in scale, offering a restricted number of videoclips, frames and identities. For instance, VioPeru  consists of only 367 clips of 46K frames, depicting only people from Peru; RWF-2000 consists of 2000 clips of 250K frames; Movies \cite{movie_fight} consists of only 200 clips of 10K frames. 
Additionally, they often lack diversity in terms of the number of individuals involved, the environments depicted, and the type of violent events. For instance, RWF-2000 contains action violence and rarely includes night scenes; VioPeru mostly features night scenes.  Hockey Fight includes only hockey-related videos depicting two-person fights (action violence) and the background is very similar in all videos.
The third is the \underline{Limited Diversity of Footage}. Current databases often focus exclusively on either surveillance or non-surveillance footage, ignoring the diverse types of footage present in real-world scenarios. For instance, VioPeru, RWF-2000  contain only surveillance videos; Hockey Fight, Movies contain videos of tv cameras for hockey games and  mostly boxing.

The fourth relates to \underline{Small and Fixed Video Lengths}. Many databases contain videos of short and fixed lengths, which fail to capture the complexity of real-world scenarios where violent incidents vary significantly in duration and temporal structure. For instance, each video in VioPeru and RWF-2000 has a fixed duration of 5 seconds; each video in Movies and Hockey Fight has a duration of $<3$ seconds. 
The fifth one is \underline{Staged Footage}. Databases such as VioShot \& Movies rely heavily on acted or movies  scenes,  lacking the spontaneity and unpredictability of real-world violence.
The sixth one is the \underline{Challenging Contexts}. 
Confusing scenarios, such as crowds of people walking or individuals engaging in non-violent actions like high-fives, are underrepresented, despite their importance in testing model robustness.
In Hockey Fight, individuals are almost always wearing the same sports clothing and the violent action occupies almost the entire frame. 
In other databases, such as VioShot and Movies, the camera adopts characteristics and positions oriented toward the best shot; however, in a real-world scenario, this does not happen.

The seventh one is the \underline{Unrealistic Balance and} \underline{Resolutions}. Many databases, such as UCF Crime and RWF-2000, are ``artificially'' balanced, containing equal numbers of violent and non-violent videos, which fails to reflect the imbalance found in real-world data. Many databases, such as Human Violence and VioShot, also predominantly feature high-resolution videos, which is not representative of practical scenarios where low-resolution footage is common.
The eight one is the \underline{Bias Against Women}. These databases predominantly focus on incidents involving men, resulting in a gender bias for the models. 
%
%
%
%
Last but not least is the  \underline{Absence of Metadata}. Contextual information, such as the number of participants involved, the type of recording equipment used and other crucial contextual attributes previously discussed, is often missing.
The absence of such metadata restricts the ability to develop models that have situational awareness, can generalize across different environments, and are more robust and explainable.

To address all above shortcomings, we present Diverse video Violence Database (DVD), a large-scale and diverse collection of 500 videos (2.7M frames) annotated at frame-level for violence detection. DVD captures high variability in real-world violence, featuring multiple scenes per video, diverse lighting conditions, varying occlusion levels, different sound environments. It includes a wide range of challenging contexts and rich metadata, detailing the number of participants (including women), type of event, scene descriptions, camera footage, and environmental factors. 
Table \ref{dbs} provides a comprehensive comparison of the features of widely used violence detection databases against those of DVD, highlighting its unique strengths and innovations.

\noindent
To sum up, this work makes the following \textbf{contribution}:
\begin{itemize}
    \item DVD Database, a large-scale, frame-level annotated violence detection database designed to address the limitations of existing databases and enhance real-world applicability. DVD will be publicly released upon acceptance;


\end{itemize}

\section{Databases}

Here, we briefly mention some widely used datasets. More details about the datasets can be found in Table \ref{dbs}.  

The Real World Fighting (RWF-2000) dataset \cite{rwf_dataset} (published in 2020) contains real world fighting scenarios in surveillance footage sourced through public platforms such as YouTube. RWF-2000 contains 2,000 trimmed video clips where each video is trimmed to 5 seconds. The dataset is  balanced with 1000 violent videos and 1000 non-violent videos, with a 80\%-20\% predefined train-test split. 

The Real Life Violence Situations (RLVS) dataset \cite{reallife_violence_dataset}  contains 2000 video clips with 1000 violent and another 1000 non-violent videos. It contains many real fight situations in several environments and conditions with an average length of 5 secs. These videos have been sourced from YouTube to include diverse scenes such  as surveillance footage, movie scenes and video recordings. A 80\%-20\% train-test split has been created for this dataset. 

As one of the latest violence detection databases published in 2024, VioPeru~\cite{baca2024} consists of 280 videos collected from real video surveillance camera records. The videos have been collected from the citizen security offices of different municipalities in Peru. It also includes 87 non-violent videos. 
The videos have been trimmed to 5 secs just to include the relevant incident. The authors in~\cite{baca2024} have described this database to contain challenging violent incidents involving two or more people captured in different environments at different times of the day using cameras with varying resolutions. 

The datasets Hockey Fight~\cite{hockeyfight} and Movies~\cite{movie_fight} were not included in the comparison as ~\cite{hockey_movie_best} reports a fairly standard multi-stream CNN has been able to achieve 100\% accuracy on the test of both the databases - so already the performance has reached the maximum in classification of these 2 databases test sets.

\section{Diverse video Violence Database (DVD)}

\subsection{Collection}
The construction of our Diverse video Violence Database (DVD) involved a meticulous and systematic data collection process aimed at ensuring diversity, inclusivity, and real-world applicability. We sourced videos from \textit{YouTube}, leveraging its extensive repository of publicly available footage. Our process included the following steps:

\noindent
\textbf{1) Keyword-Based Search:} We specified a comprehensive set of keywords related to violence to guide our search. These keywords were designed to capture a wide range of violent scenarios, actions, people and contexts, including:
\begin{itemize}
    \item Types of violence and actions (e.g., street fights, bar brawls, domestic disputes, crowd violence, knife attacks, gunfire incidents, arrests turned violent, riot clashes, punching, kicking, pushing);
     \item Locations of violence (e.g., public spaces, indoor venues, outdoor markets, residential areas, workplaces);
     \item Actions leading to violence (e.g., arguments escalating into fights, property damage followed by physical violence, self-defense incidents);
     \item Crowd dynamics (e.g., large groups interacting, individual altercations, protests turning violent);
     \item Women: videos featuring women as participants in violent (e.g. as victims, or abusers) or non-violent events;
     \item Diverse footage (e.g., mobile phone recordings, CCTV, body cameras, dashcams, in-vehicle cameras).
\end{itemize}

\noindent
\textbf{2) Multilingual Search:} 
To enhance diversity and ensure representation of various cultures and settings, we repeated the keyword search in 6 different languages, including English, Spanish, German, French, Chinese, and Hindi. In this way we captured videos featuring individuals of various nationalities, ensuring a global perspective on violent events.

\noindent
\textbf{3) Quality Control and Filtering:} After the initial collection, we manually reviewed the videos to ensure relevance and clarity. We removed ambiguous content; videos in which the violent anomaly was unclear or poorly visible were discarded. We also ensured real-world applicability; videos that appeared heavily staged, acted, or unrelated to violence detection tasks were excluded.

\subsection{Annotation}

To ensure high-quality annotations, four annotators were selected to perform the annotation task. These annotators were computer scientists with a foundational understanding of violence detection. To enhance their annotation skills and ensure suitability for the task, they underwent additional training, including practical exercises and detailed guidelines tailored to the unique aspects of the database. Each annotator was instructed both orally and through a comprehensive multipage document that outlined the procedure for annotating the videos. This document included:

\begin{figure*}[ht]
  \centering
   \includegraphics[width=1.\linewidth]{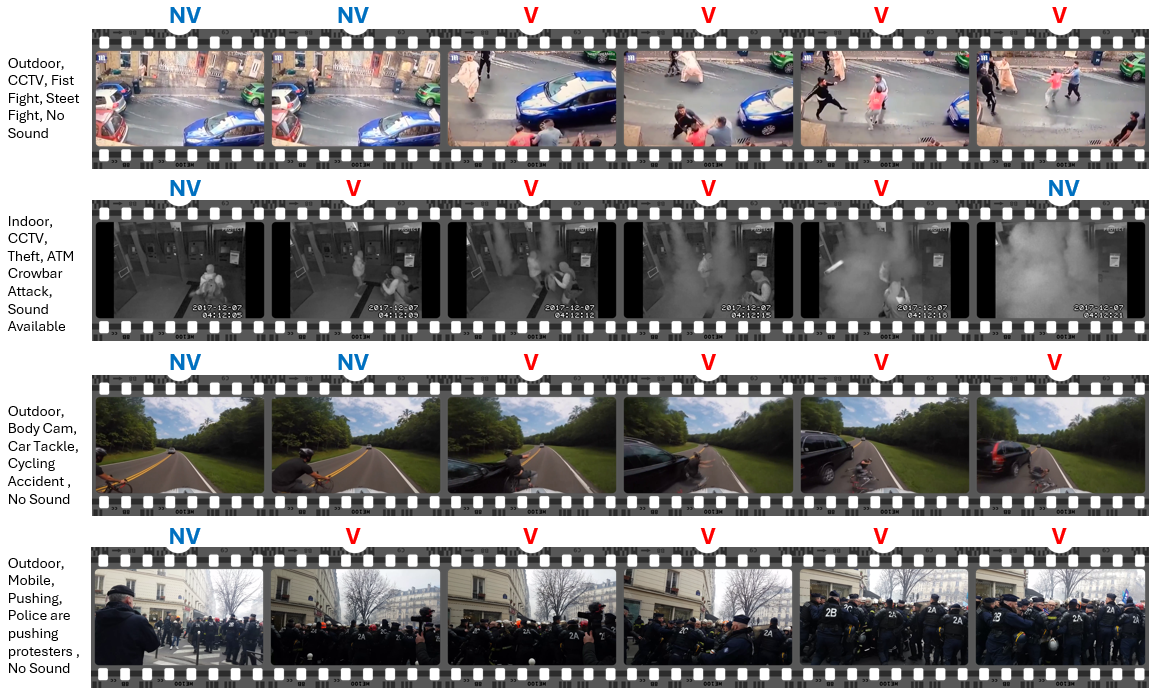}
   \caption{Example videos from our DVD database, showcasing diverse scenes, environments, participants, and camera types. Each video in DVD database includes granular labeling of violent frames and annotated metadata for detailed analysis.}
   \label{fig222}
\end{figure*}

\begin{itemize}
    \item Detailed explanations of what constitutes a violent event, with examples of well-defined scenarios (e.g., physical altercations, crowd violence, weapon-based incidents), as well as what types of violence exist;
  \item Detailed explanations of how to annotate the violence events: the labeled segments should include a few additional frames/seconds at the start and end of each violent event to provide context. Additionally, brief pauses during a violent event, where no strikes or hits occur but individuals remain in a confrontational or aggressive position, should be labeled positive. However, extended breaks, where the aggressive behavior ceases and the situation de-escalates, should be labeled as negative. Any subsequent violent activity after such a break should be considered a new and separate instance of violence;
    \item Guidance on handling scenarios such as non-violent interactions that could resemble violence (e.g., handshakes or playful gestures);
    \item Instructions on identifying and annotating metadata.
\end{itemize}

Before initiating the main annotation task, we conducted an initial round of testing to ensure the annotators fully understood the process. During this phase, the annotators: i) annotated a small subset of videos individually; ii) participated in cross-annotation checks, where the annotations of each annotator were compared for consistency and alignment with the guidelines; iii) received feedback and clarification on discrepancies to improve their accuracy and adherence to the annotation protocol.
This preparatory phase was crucial for achieving a high degree of inter-annotator agreement and ensuring consistency across the database.

Before starting the annotation of each video,
the experts watched the whole video so as to know what to expect regarding the violence events taking place in the video. Finally, the annotators annotated all videos.
\underline{Annotation Post-Processing}
To further validate the annotations, a post-processing step was performed. In this phase each annotator reviewed their initial annotations by watching the videos a second time. They verified that their recorded annotations accurately reflected the events depicted in the videos. Adjustments were made where necessary, particularly in cases of ambiguous or borderline scenarios. A cross-annotator validation step was conducted, where all four annotators reviewed a subset of videos annotated by their peers. This helped identify potential inconsistencies and ensured a consensus on complex cases. Once all annotations were finalized, we kept only the annotations on which at least three experts agreed. A final quality check was performed by an independent reviewer to ensure the annotations adhered to the established guidelines. This reviewer focused on the clarity and accuracy of the binary violence labels for each frame, as well as on the completeness, consistency and correctness of metadata annotations \cite{kollias2024sam2clip2sam,zafeiriou1,Tailor,springer,kollias20247th,kollias2023multi,kollias2024distribution,psaroudakis2022mixaugment,kollias2021distribution,kollias2019face,kollias2019deep,kollias2019expression,kollias2022abaw,kollias2023abaww,kollias2023btdnet,kollias2023facernet,kollias2023multi,kollias20246th,kollias2024distribution,kolliasijcv,zafeiriou2017aff,hu2024bridging,psaroudakis2022mixaugment,kollias2020analysing,kollias2021distribution,kollias2021affect,kollias2019face,kollias2021analysing,kollias2023ai,kollias2023deep2,arsenos2023data,salpea2022medical,arsenos2024uncertainty,karampinis2024ensuring,arsenos4674579nefeli,miah2024can,arsenos2024commonn,cis,hu2024rethinking,gerogiannis2024covid,kollias2024behaviour4all,kollias2024domain,kollias2020va,kollias2023mma,kollias6,kollias2018old,kollias2018photorealistic,kollias2017adaptation,tagaris1,tagaris2,kollias2016line,kollias2015interweaving,Kollias2025,kolliasadvancements,arsenos2024common22,arsenos2022large,ref100,kollias2020exploiting,kollias2018multi,kollias2018aff,arsenos2024common}. 


\subsection{Database Properties \& Examples}

\begin{figure}[h]
  \centering
   \includegraphics[width=1.\linewidth]{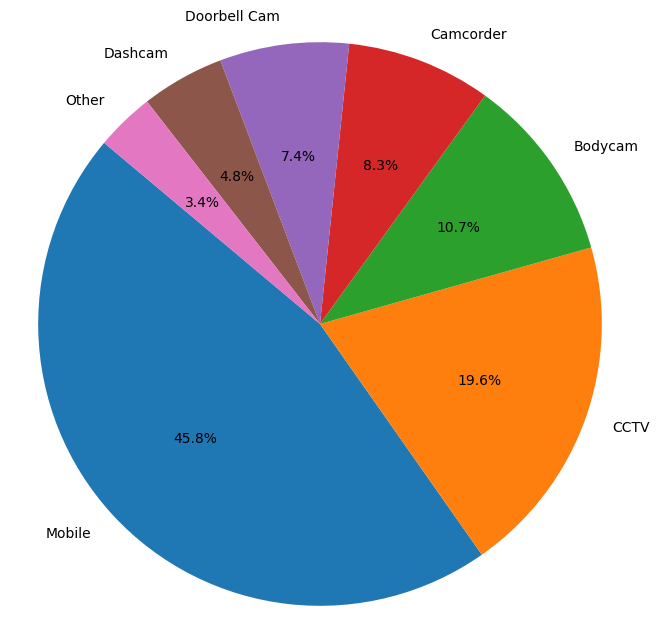}
   \caption{Distribution of video footage types in the database, with each slice representing the percentage of videos recorded using a specific camera type.}
   \label{fig22}
\end{figure}

Figure \ref{fig222} shows some example videos from DVD with wide range of scenes, environments, participants and camera types, along with granular labeling (indicating which frames have violence) and annotated metadata.
Table \ref{dbs1} illustrates how many outdoors and indoor scenes are included in DVD, as well as in how many cases sound was/wasn't linked to the violent/non-violent event. 
Figure \ref{fig22} illustrates the distribution of different video footage in DVD. 
The lowest video resolution is $320 \times 240$ and the highest is $3840 \times 2160$. Finally, DVD has been split into training, validation and test sets in 55-15-30\% ratio, maintaining  similar label distributions; the database partitioning has been manually verified for any form of data leakage. 

 \begin{table}[h]
 \centering
 \caption{Some properties of the proposed DVD database.
 }
 \scalebox{.9}{
 \begin{tabular}{ll|ll}
 \toprule
 \textbf{Scene Type} & \textbf{Num} &  \textbf{Sound Association} & \textbf{Num} \\
 \toprule
   Outdoor Scenes & 1457 &  Has sound linked to event & 1263 \\ 
    Indoor Scenes & 543 & No sound linked to event & 737\\ 
 \bottomrule
 \end{tabular}
 }
 \label{dbs1}
 \end{table}

\noindent
\textbf{DVD Database's Novelties}
The novelties of our database are summarized below:
\begin{itemize}

\item \textbf{Scale, Video Lengths \& Diversity}: DVD consists of 500 long videos (of variable lengths, from 15 to 1200 seconds) of around 2.7 million frames. The videos exhibit: i) variable resolutions; ii) multiple scenes per video, diverse illumination conditions (day/night), varying levels of occlusion, and distinct sound levels; iii) 
individuals from different nationalities involved in the events; iv)  variety of footages; v) different types of violence; vi) challenging contexts (e.g., crowded environments where large groups of people interact or walk in close proximity; partial occlusion of violent scenes by other people or objects; non-violent actions like high-fives, handshakes, or hugging that may appear ambiguous; complex backgrounds, such as busy streets, markets, or heavy traffic; cases where violent incidents are either small or large relative to the size of the video frame; individuals wearing masks, helmets, or hats, obscuring key features; background dynamics, including cheering, clapping, jumping, or bystanders reacting emotionally by running or screaming). 

\item \textbf{Frame-Level Annotations}: each video contains  violent and non-violent frames; the annotation is at frame-level.


\item \textbf{Inclusion of Women}: DVD contains 200 videos that include women in violent events (either as victim or perpetrator) and non-violent events. 


\item \textbf{Rich Per-Frame Metadata}: We provide detailed metadata for each frame, including: i) number of people involved; ii) type of the event, such as bar fights, arrests, knife attacks, and gunfire incidents; iii) a description characterizing the event or its context, such as `helicopter video captures shootout on highway', `police bodycam footage shows intense shootout with suspect', `gas station armed robbery', `cheerleading', `stadium collapses while fans celebrate victory';  iv) type of footage, such as mobile phones, surveillance cameras, body cameras, dashcams, camcorders, doorbell cameras, and in-vehicle cameras; v) whether the scene is indoors or outdoors; vi) whether there is sound associated with the violent event.


\end{itemize}

\subsection{Images}
\begin{figure*}[htp]
  \centering
   \includegraphics[width=1.\linewidth]{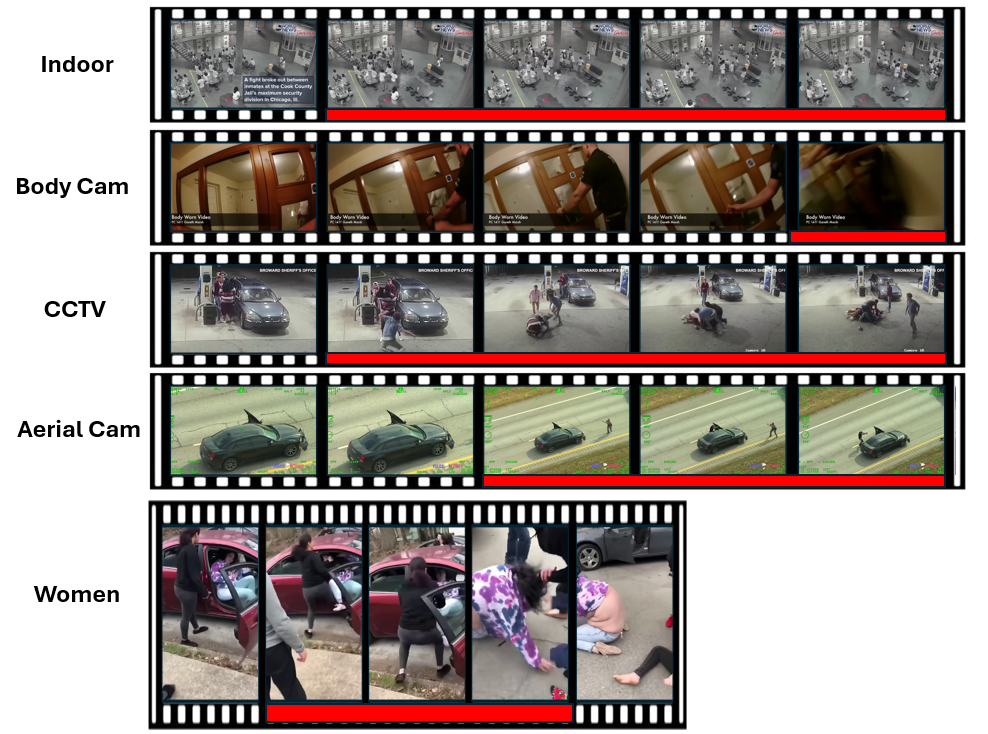}
   \caption{Example videos from the Diverse video Violence Database (DVD).  The database consists of videos from a range of scenes, environments, participants and camera types, along with granular labeling (visualized in red) indicating which frames have violence.}
   \label{fig:videos}
\end{figure*}

Figure \ref{fig:videos} shows some example videos from DVD with a wide range of scenes, environments, participants and camera types, along with granular labeling (visualized in red) indicating which frames have violence.

\subsection{Label Imbalance}
While DVD database contains more non-violent frames than violent ones, this design reflects real-world conditions where violent events are typically rare. 
The difference between the number of violent and non-violent frames could make people think that it could lead to train a biased model and produce wrong predictions.
To avoid this issue, during model training,  we  employ weighted loss function to prevent the model from being skewed toward the dominant class. Our experimental results demonstrate that models trained on DVD generalize well across the same database or across other databases, indicating that the database's distribution does not negatively impact model performance.

\subsection{Collection}
\paragraph{Keyword-Based Search}

Keywords were initially searched individually, and in later stages, we experimented with keyword combinations to refine search results. Combinations were generated based on logical groupings, such as pairing “street fights” with “public spaces” or “gunfire incidents” with “police bodycam.”

\paragraph{Multilingual Search.}
To ensure diverse cultural representation, we conducted multilingual searches using keywords in 6 different languages, including English, Spanish, German, French, Chinese, Hindi. Each translation was carefully reviewed to account for contextual differences in how violence is described across regions. We cross-validated retrieved videos to ensure consistency in violence interpretation, filtering out cases where cultural discrepancies led to ambiguous categorization. This process enhances the database’s applicability in global research settings.

\paragraph{Database Properties.}
Table 2  illustrates how many outdoors and indoor scenes are included in DVD, as well as in how many cases sound was or was not linked to the violent or non-violent event. Let us mention that no sound linked to the event, can either mean that there is no audio recorded or that there is audio but it is irrelevant to the violent or non-violent event.  One can see that in most of the cases, sound is linked to the violent or non-violent event and our database contains many cases with indoor scenes (and many more with outdoor scenes).

\begin{figure*}[t]
  \centering
   \includegraphics[width=1.\linewidth]{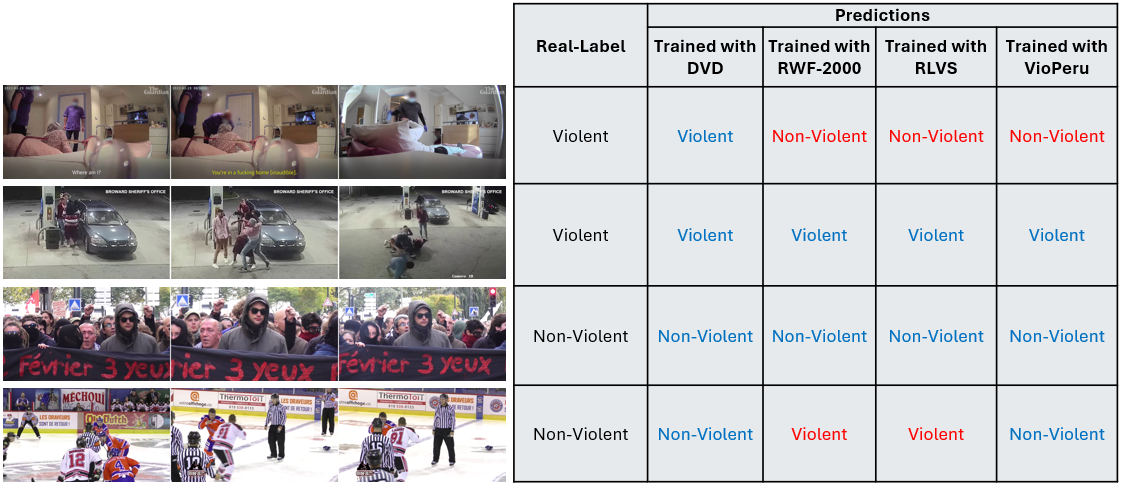}
   \caption{DVD video samples showcasing its labels and the predictions of CLIP-VDNet trained on DVD, RWF-2000, RLVS, and Vio Peru.}
   \label{aaa}
\end{figure*}

Finally, Figure \ref{aaa} shows a few videos in DVD, along with their corresponding ground truth labels and predictions made by a network \cite{chamalke2024cue,senadeera2024cue} when trained on DVD, RWF-2000, RLVS and  Vio Peru.

\section{Conclusion}

In this paper, we introduced DVD, a large-scale and diverse violence detection database designed to address the limitations of existing databases and advance research in this critical domain. By incorporating frame-level annotations, detailed metadata, and a wide range of scenarios, environments, and participants —including women —, our database sets a new standard for comprehensiveness and inclusivity. 

By making DVD publicly available, we aim to foster innovation and collaboration within the research community, enabling the development of more accurate, reliable, and equitable solutions for violence detection in real-world settings. 


{
    \small
    \bibliographystyle{ieeenat_fullname}
    \bibliography{main}
}

\end{document}